\UseRawInputEncoding
\documentclass{article}
\usepackage{spconf,amsmath,graphicx}
\usepackage{booktabs}
\usepackage{threeparttable}
\usepackage{multirow}
\usepackage{amsfonts}
\usepackage{amsmath}
\usepackage{mathrsfs}
\usepackage[ruled]{algorithm2e}
\DeclareMathSizes{10}{10}{5}{5}

\title{Edge-Labeling based Directed Gated Graph Network for Few-shot Learning}
\name{Peixiao Zheng, Xin Guo$^{*}$, Lin Qi\thanks{*Corresponding author: Xin Guo, iexguo@zzu.edu.cn}}
\address{School of Information Engineering, Zhengzhou University, Zhengzhou, China}
\begin{document}
%
\maketitle
\begin{abstract}
Existing graph-network-based few-shot learning methods obtain similarity between nodes through a convolution neural network (CNN). However, the CNN is designed for image data with spatial information rather than vector form node feature. In this paper, we proposed an edge-labeling-based directed gated graph network (DGGN) for few-shot learning, which utilizes gated recurrent units to implicitly update the similarity between nodes. DGGN is composed of a gated node aggregation module and an improved gated recurrent unit (GRU) based edge update module. Specifically, the node update module adopts a gate mechanism using activation of edge feature, making a learnable node aggregation process. Besides, improved GRU cells are employed in the edge update procedure to compute the similarity between nodes. Further, this mechanism is beneficial to gradient backpropagation through the GRU sequence across layers. Experiment results conducted on two benchmark datasets show that our DGGN achieves a comparable performance to the-state-of-art methods.
\end{abstract}
\section{Introduction}
\label{sec:intro}

Deep neural networks \cite{Krizhevsky2017ImageNet,2017Revisiting} based on massive labeled data have achieved great success in recent years. However, the process of obtaining labeled data is cumbersome. Therefore, training a robust model with a small amount of labeled data is an urgent issue. Few-shot learning \cite{2006One,lake2011one} aims to predict unlabeled data (query set) based on a few labeled data (support set).

There are three main ways to address the few-shot learning problem. The first category of few-shot learning approaches adopts a metric learning framework to minimize the distance between the same class samples. Vinyals \emph{et al}. \cite{vinyals2016matching} assumed a weighted nearest neighbor classifier using an attention mechanism. Snell \emph{et al}. \cite{snell2017prototypical} produced a prototype embedding through the average of each class feature. Sung \emph{et al}. \cite{sung2018learning} built a distance metric network to obtain point-wise relations in all samples. The second category of few-shot approaches focused on extracting transferable knowledge across tasks. Finn \emph{et al}. \cite{finn2017model} aimed to find the best set of initialization parameters that will allow it to achieve good performance with a few times gradient updates on any new task. By simplifying \cite{finn2017model}, Nichol \emph{et al}. \cite{nichol2018first} was a first-order gradient-based meta-learning algorithm.

Since the above two kinds of methods have shown the requirements of the relationship between a support set and a query set, the third category of few-shot learning approaches applied graph neural network (GNN) \cite{scarselli2008graph,kipf2016semi} to further excavate the implicit relational information between samples. Garcia \emph{et al}. \cite{garcia2017few} utilized annotation information to initialize the adjacency matrix and then updates the node information in the graph continuously through the message passing process. Liu \emph{et al}. \cite{liu2018learning} firstly adopted a transductive setting on graph-based few-shot learning, which using a CNN module to compute the similarity between nodes and propagate labels from support set to query set in the graph. To further exploit the intra-cluster similarity and inter-cluster dissimilarity of the nodes in graph neural network, Kim \emph{et al}. \cite{kim2019edge} built an edge-labeling graph neural network framework, which performs well by alternating node and edge feature updates. However, it is not reasonable enough to use a CNN to measure the similarity of vector format node information in the above methods \cite{liu2018learning,kim2019edge}, and the node feature update process \cite{kim2019edge} is also an unlearnable node feature sum procedure.

This paper propose a novel directed gated graph neural network (DGGN) based on edge annotation for few-shot learning. As shown in Figure 1, the framework of DGGN is composed of two parts: Node update and Edge update. The node update process adopts a gate mechanism using the activation of edge feature to control the aggregation of node feature. Such a learnable aggregation approach can easily  incorporate into an end-end network. To obtain the similarity between the nodes expressed as vectors, we propose an improved GRU \cite{chung2014empirical} mechanism instead of CNN.
\begin{figure*}[ht]
\begin{minipage}[b]{1.0\linewidth}
  \centering
  \centerline{\includegraphics[width=16cm]{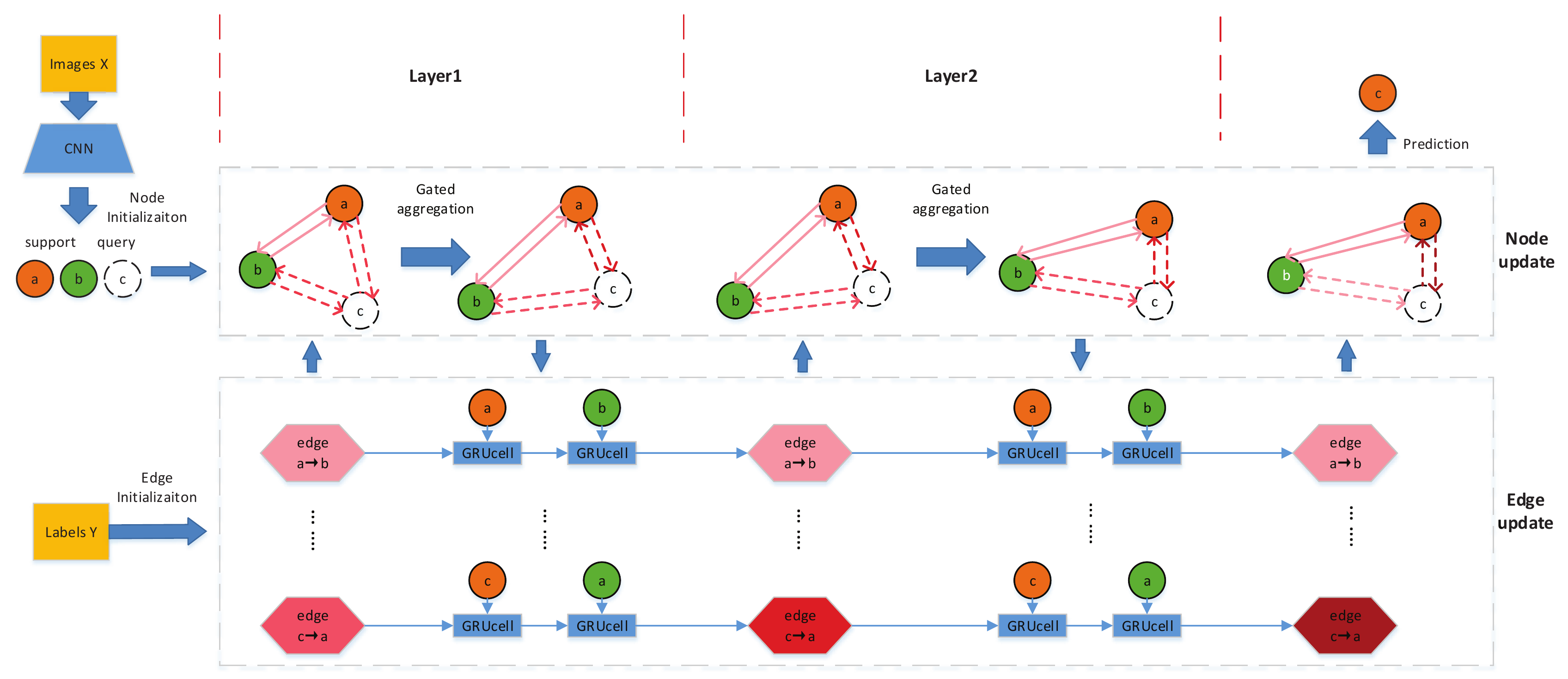}}
\end{minipage}

\caption{The framework of a two layers DGGN. The node update and edge update are carried out alternately, and the class prediction of nodes is made based on the final edge features. Please refer to the text for more details.}
\label{fig:framework}
\end{figure*}
Expressly, we set the edge feature as the hidden state and the node feature as the input. Then, the features of the two nodes connected by the directed edge are input into the GRU in the edge direction. Thus, the edge feature as the hidden state of the GRU is updated. After a few DGGN layers conducted by these two mechanisms, nodes in the graph can be classified by simple weighted voting based on edge feature. Our contributions can be summarized as follows:
\begin{itemize}
\item The gated node update mechanism makes the node feature aggregation a learnable process, which can easily incorporate into an end-end framework.
\item The proposed improved GRU mechanism is appropriate for vector form node features compared with CNNs, and the GRU sequence throughout each layer is benefit for gradient backpropagation.
\item On few-shot image classification tasks, the proposed method providing improved numerical performance on two benchmark datasets.
\end{itemize}
\section{Method}
\subsection{Problem definition:Few-shot classification}
\label{ssec:def}
The goal of few-shot classification is to construct a robust classifier when only a few labeled training samples are given. Each few-shot classification task $\mathcal{T}$ contains a support set $\mathcal{S}$ and a query set $\mathcal{Q}$. If the support set $\mathcal{S}$ contains K labeled samples for each N class, it is called N-way K-shot classification. Specifically, given a training datasets $\mathbb{D}$$^{train}$ and a testing datasets $\mathbb{D}$$^{test}$, although both $\mathbb{D}$$^{train}$ and $\mathbb{D}$$^{test}$ are sampled from one datasets $\mathbb{D}$, which means they have similar distribution,the labels of $\mathbb{D}^{train}$ and $\mathbb{D}$$^{test}$ are mutually exclusive. In the training stage, many N-way K-shot tasks are built on $\mathbb{D}$$^{train}$ as follows:
\begin{equation}\mathcal{S}=\{(x_1,y_1),(x_2,y_2),...,(x_{N\times{K}},y_{N\times{K}})\},\end{equation}
\begin{equation}\mathcal{Q}=\{(x_{N\times{X}+1},y_{N\times{X}+1}),...,(x_{N\times{X}+C},y_{N\times{X}+C})\},\end{equation}
where $x_i$ is the samples in $\mathbb{D}$$^{train}$, $y_i$ is the corresponding label of $x_i$, C is the number of query samples. Then, the support set $\mathcal{S}$ in each task is taken as a training set on which the model is trained to minimize the loss of its predictions over the query set $\mathcal{Q}$. As for the testing, sampling the support sets and query sets as the same way as the training stage and employing the trained model to predict the query sets.The labels of the support set and test set in the training phase are both provided for model optimization, while only the support set‘s labels are provided in the test phase.This kind of training procedure is called episode training \cite{vinyals2016matching,santoro2016meta}.
\subsection{Node and edge initialization process}
\label{ssec:initial}
This section describes the node and edge initialization process of the proposed DGGN. As shown in figure 1, all samples in support set $\mathcal{S}$ and query set $\mathcal{Q}$ are embedded into feature vectors through a CNN module. Thus, a directed graph is initialized where each node represents each sample, and each edge represents the relationship between the two connected nodes. Here, $\mathcal{G}=(\mathcal{V,E;T})$ be the graph build by samples of task $\mathcal{T}$, where $\mathcal{V}= \{V_i\}_{i=1,...,T}$ denote the nodes of the graph and $\mathcal{E}= \{E_{ij}\}_{i,j=1,...,T}$ denote the edges, $T=N\times K+C$ is the number of all samples in the task $\mathcal{T}$. $\boldsymbol v_i$ is the feature of node $V_i$ and $\boldsymbol e_{ij}$ is the feature of edge $E_{ij}$. Furthermore, we can define the ground-truth edge label $y_{ij}$ by node labels $y_i$ as follows:
\begin{equation}
y_{ij}=\left\{
\begin{aligned}
1 & , & y_i=y_j, \\
0 & , & y_i\ne y_j.
\end{aligned}
\right.\end{equation}

Node feature is initialized by a convolutional neural network: $\boldsymbol v_i^{\ell=0}=f_{cnn}(x_i;\theta_{cnn})$, where $f_{cnn}$ is the parameter set of the CNN and $\ell$ is the number of layer. We use the CNN with four convolution blocks for a fair comparison, which is employed in most few-shot learning models \cite{snell2017prototypical,vinyals2016matching,garcia2017few,liu2018learning,kim2019edge}. Specifically, each convolution block contains $3\times 3$ size kernels, batch normalization \cite{ioffe2015batch} and a ReLU activation module. Each edge feature $\boldsymbol e_{ij} = (e_{ij1},e_{ij2})$ is a two dimensional vector, which represent the intra-class similarity and inter-class dissimilarity of the two connected nodes separately. Then, the edge features are initialized by edge label $y_{ij}$ as follows:
\begin{equation}\boldsymbol e_{ij}^{\ell=0}=\left\{
\begin{aligned}
(1&,0),  \qquad y_{ij}=1  \quad and  \quad i,j\leq N\times K,  \\
(0&,1),  \qquad y_{ij}=0  \quad and  \quad i,j\leq N\times K,  \\
(0.5&,0.5),   \qquad \qquad otherwise.
\end{aligned}
\right.\end{equation}
\subsection{Gated node update mechanism}
\label{ssec:node}
Given node feature $\boldsymbol v_i^{\ell}$ and edge feature $\boldsymbol e_{ij}^{\ell}$ of layer $\ell$, node update is firstly conducted based on a gated aggregating mechanism. Inspired by the original edge gating method \cite{marcheggiani2017encoding}, our node update procedure is designed as follows:
\begin{equation}\boldsymbol v_i^{\ell+1} = ReLU(A^{\ell}\boldsymbol v_i^{\ell} + \sum\limits_{j\to i}\sigma(C^{\ell}\boldsymbol e_{ij}^{\ell})\odot B^{\ell }\boldsymbol v_j^{\ell}),\end{equation}
where $\sigma$ is the sigmoid function, $A, B, C$ are the weight parameters and $\odot$ means element-wise product. Different from \cite{kim2019edge}, we utilize the activation of edge feature as a gate to control the message passing from neighbor nodes $\boldsymbol v_j$ to the center node $\boldsymbol v_i$. Therefore, the overall aggregating process is becoming learnable compared with the fixed node updating process in EGNN.
\subsection{GRU based edge update mechanism}
\label{ssec:edge}
To update the edge feature properly on the condition that node features are vector formed, we construct a GRU sequence-based approach as follows:
\begin{equation}
\boldsymbol e_{ij}^{\ell+1} = GRUseq(\boldsymbol e_{ij}^{\ell}, \boldsymbol v_i^{\ell}, \boldsymbol v_j^{\ell}),
\end{equation}
\begin{small}
\begin{equation}
GRUseq(\boldsymbol e_{ij}^{\ell}, \boldsymbol v_i^{\ell}, \boldsymbol v_j^{\ell}) = GRU_2( GRU_1(\boldsymbol e_{ij}^{\ell}, \boldsymbol v_i^{\ell}), \boldsymbol v_j^{\ell}),
\end{equation}
\end{small}where $\boldsymbol e_{ij}^{\ell}$ the directed edge connecting the nodes from $\boldsymbol v_i^{\ell}$ to $\boldsymbol v_j^{\ell}$, and $GRU$ represents the gated recurrent unit\cite{chung2014empirical} widely used for natural language processing tasks \cite{li2015gated}. More concretely, $GRU_1(\boldsymbol e_{ij}^{\ell}, \boldsymbol v_i^{\ell})$ is equal to:
\begin{equation}
z_i = \sigma(U_z\boldsymbol e_{ij}^{\ell}+V_z\boldsymbol v_i^{\ell}),
\end{equation}
\begin{equation}
r_i = \sigma(U_r\boldsymbol e_{ij}^{\ell}+V_r\boldsymbol v_i^{\ell}),
\end{equation}
\begin{equation}
\boldsymbol{\tilde{e}}_{ij}^{\ell} = tanh(U_e(\boldsymbol e_{ij}^{\ell}\odot r_i)+V_e\boldsymbol v_i^{\ell}),
\end{equation}
\begin{equation}
\boldsymbol{\hat{e}}_{ij}^{\ell} = (1-z_i)\odot \boldsymbol e_{ij}^{\ell}+ z_i\odot \boldsymbol{\tilde{e}}_{ij}^{\ell},
\end{equation}
where $\boldsymbol{\hat{e}}_{ij}^{\ell}$ is the updated $\boldsymbol{e}_{ij}^{\ell}$ by the GRU cell. We treat the edge feature $\boldsymbol e_{ij}^{\ell}$ as the hidden state, and update it by feeding the node features into the GRU. After two-node features are feeded into the GRU sequence in the order of edge direction, the edge feature $\boldsymbol e_{ij}^{\ell}$ can extract information from these two nodes and the relationship between the two connected nodes is obtained.

Also, we insert a residual block \cite{he2016deep} between layers to ease the network degradation situation:
\begin{equation}
\boldsymbol v_{i}^{\ell+1} = f^{\ell}_v(\boldsymbol v_{i}^{\ell}, \{\boldsymbol v_{j}^{\ell}: j \rightarrow i\}, \boldsymbol e_{ij}^{\ell})+\boldsymbol v_{i}^{\ell},
\end{equation}
\begin{equation}
\boldsymbol e_{ij}^{\ell+1} = f^{\ell}_e(\boldsymbol e_{ij}^{\ell}, \boldsymbol v_{i}^{\ell}, \boldsymbol v_{j}^{\ell})+\boldsymbol e_{ij}^{\ell}.
\end{equation}
\begin{algorithm}[ht]
\caption{The process of DGGN for inference}
\LinesNumbered 
\KwIn{$\mathcal{G}=(\mathcal{V,E;T})$, where $\mathcal{T = S \cup Q}$, $\mathcal{S}= \{(x_i,y_i)\}^{N\times K}_{i=1}$, $\mathcal{Q} = \{x_i\}_{i=N\times K+1}^{N\times K+C}$}
\KwOut{$\{\hat{y_i}\}^{N\times K+C}_{i=N\times K+1}$}
\textbf{Initialize:} $\boldsymbol v_i^{\ell=0} =  f_{cnn}(x_i;\theta_{cnn}), \boldsymbol e_{ij}^{\ell=0}, \forall i,j$\\ 
\For{$\ell = 0,..., L-1$}{
\tcc{Node feature update}

　　\For{$i = 1,..., \lvert V \rvert$}{
$\boldsymbol v_i^{\ell+1} = ReLU(A^{\ell}\boldsymbol v_i^{\ell} + \sum\limits_{j\to i}\sigma(C^{\ell}\boldsymbol e_{ij}^{\ell})\odot B^{\ell }\boldsymbol v_j^{\ell})$,}
\tcc{Edge feature update}

   \For{$(i, j) = 1,..., \lvert E \rvert$}{
$\boldsymbol e_{ij}^{\ell+1} = GRUseq(\boldsymbol e_{ij}^{\ell}, \boldsymbol v_i^{\ell}, \boldsymbol v_j^{\ell})$, }
}
\tcc{Query node label prediction}

$\{\hat{y_i}\}^{N\times K+C}_{i=N\times K+1} \leftarrow$ WeightedVoting$(\{y_i\}^{N\times K}_{i=1}, \{\boldsymbol e_{ij}^L\})$
\end{algorithm}
\subsection{Objective}
\label{ssec:obj}
After $L$ layers of alternative node and edge updates, the node prediction can be obtained from the final edge features: $\boldsymbol e_{ij}^{L}$. As we mentioned in section 2.2, $\boldsymbol e_{ij} = (e_{ij1},e_{ij2})$ represents the intra-class similarity and inter-class dissimilarity of the two connected nodes separately so that the nodes can be classified by simple weighted voting with support node's label and final edge feature $e_{ij1}^{L}$:

\begin{equation}
P(\hat{y}_i=\mathcal C_k\vert \mathcal{T}) = Softmax(\sum\limits_{\{j:j\rightarrow i\}}e_{ij1}^{L} \delta(y_j=\mathcal C_k)),
\end{equation}
where $P(\hat{y}_i=\mathcal C_k\vert \mathcal{T})$ is the probability that node $V_i$ is belong to class $\mathcal{C}_k$ and $y_j$ is the label of neighbor nodes connecting to node $\boldsymbol v_i$ in the support set.

We adopt the binary cross-entropy loss as the DGGN's loss function to minimize the differences between the output edge feature and edge label. The overall procedure is shown in Algorithm 1.

\section{EXPERIMENTS}
\label{sec:pagestyle}
\subsection{Datasets and Setups}
\label{ssec:datasets}
We evaluated our DGGN on two standard few-shot learning datasets: miniImagenet \cite{vinyals2016matching} and tieredImagenet \cite{ren2018meta}.
The miniImageNet and tieredImageNet are the subsets of ImageNet \cite{russakovsky2015imagenet}. The miniImagenet contains 100 classes with 600 images per class, which are randomly split into 64, 16, 20 classes as training, validation and testing set,respectively. The tieredImagenet has 608 classes divided into 351, 97, 160 classes as training, validation and testing set. The average number of images per category in tieredImagenet was 1281.

We conducted the standard 5-way 5-shot and 5-way 1-shot experiments on these two datasets. Considering the effectiveness of the feature updates and the avoidance of over-smoothing issues, we adopt a three-layer DGGN model. The proposed DGGN model was trained with Adam optimizer and the initial learning rate was set to 0.001. Besides, We decay the learning rate by 0.1 per 20000 iterations and set the weight decay to $10^{-6}$. Our code\footnote{The DGGN code is available on https://github.com/zpx16900/DGGN.} was developed in Pytorch \cite{paszke2017automatic} framework and run with NVIDIA Tesla V100.
\begin{table}[ht]
  \begin{threeparttable}
  \caption{Few-shot classification performance on miniImagenet and tieredImagenet. Top results are highlighted.}
  \setlength{\tabcolsep}{2.5mm}
  \label{tab:tabel1}
    \begin{tabular}{ccccccc}
    \toprule
    \multirow{3}{*}{Method}&
    \multicolumn{2}{c}{miniImagenet}&\multicolumn{2}{c}{tieredImagenet}\cr
    &\multicolumn{2}{c}{5-way}&\multicolumn{2}{c}{5-way}\cr
    \cmidrule(lr){2-3} \cmidrule(lr){4-5}
    &1-shot&5-shot&1-shot&5-shot\cr
    \midrule
    MatchingNet \cite{vinyals2016matching}&43.56&55.31&-&-\cr
    ProtoNet \cite{snell2017prototypical}&49.42&68.2&53.34&72.69\cr
    RelationNet \cite{sung2018learning}&50.44&65.32&54.48&71.32\cr
    MAML \cite{finn2017model}&48.70&55.31&51.67&70.30\cr
    GNN \cite{garcia2017few}&50.33&66.41&-&-\cr
    TPN \cite{liu2018learning}&55.51&69.84&59.91&73.30\cr
    EGNN \cite{kim2019edge}&59.63&76.34&{63.52}&80.24\cr
    \textbf{DGGN}&{\bf 60.95}&{\bf 78.04}&\bf{63.98}&{\bf 81.16}\cr
    \bottomrule
    \end{tabular}
    \end{threeparttable}
\end{table}
\subsection{Experiment Results}
\label{ssec:results}
We compared our approach with several state-of-the-art methods, including graph-based and non-graph-based methods. For fair comparisons, we evaluate DGGN on miniImagenet and tieredImagenet, which is compared with other
methods in the same ConvNet4 backbone. For miniImagenet, the number of iterations is 100k with batch size = 20. In contrast, for tieredImagenet, the iteration is doubled to 200k because it is a larger dataset and needs more iterations to make the model converge. As shown in Table 1, the proposed DGGN model outperforms other existing methods and achieves state-of-the-art performance on both 5way-5shot and 5way-1shot settings.

As can be seen from Table 1, graph-based method(e.g. EGNN,DGGN) is much powerful than non-graph-based approach(e.g. ProtoNet,MAML). The results demonstrate that graph neural network is appropriate for processing relation information. Notably, DGGN achieves higher accuracies than EGGN, which adopts a CNN module to calculate the similarities between nodes to update the edge feature. In contrast, our DGGN employs a GRU sequence mechanism to process the vector form node features, which is more proper than CNN.

The performance of DGGN in 10way-5shot, 10way-1shot, and 20way-1shot situations are evaluated on miniImageNet dataset, its results are shown in Figure 2. It should be noted that our DGGN performs better than other methods in high way scenarios. With the increasing number of support sets, DGGN can build a bigger graph to conduct the node and edge updating procedure and achieve higher accuracy.

\begin{figure}[t]

\begin{minipage}[b]{1.0\linewidth}
  \centering
  \centerline{\includegraphics[width=9cm]{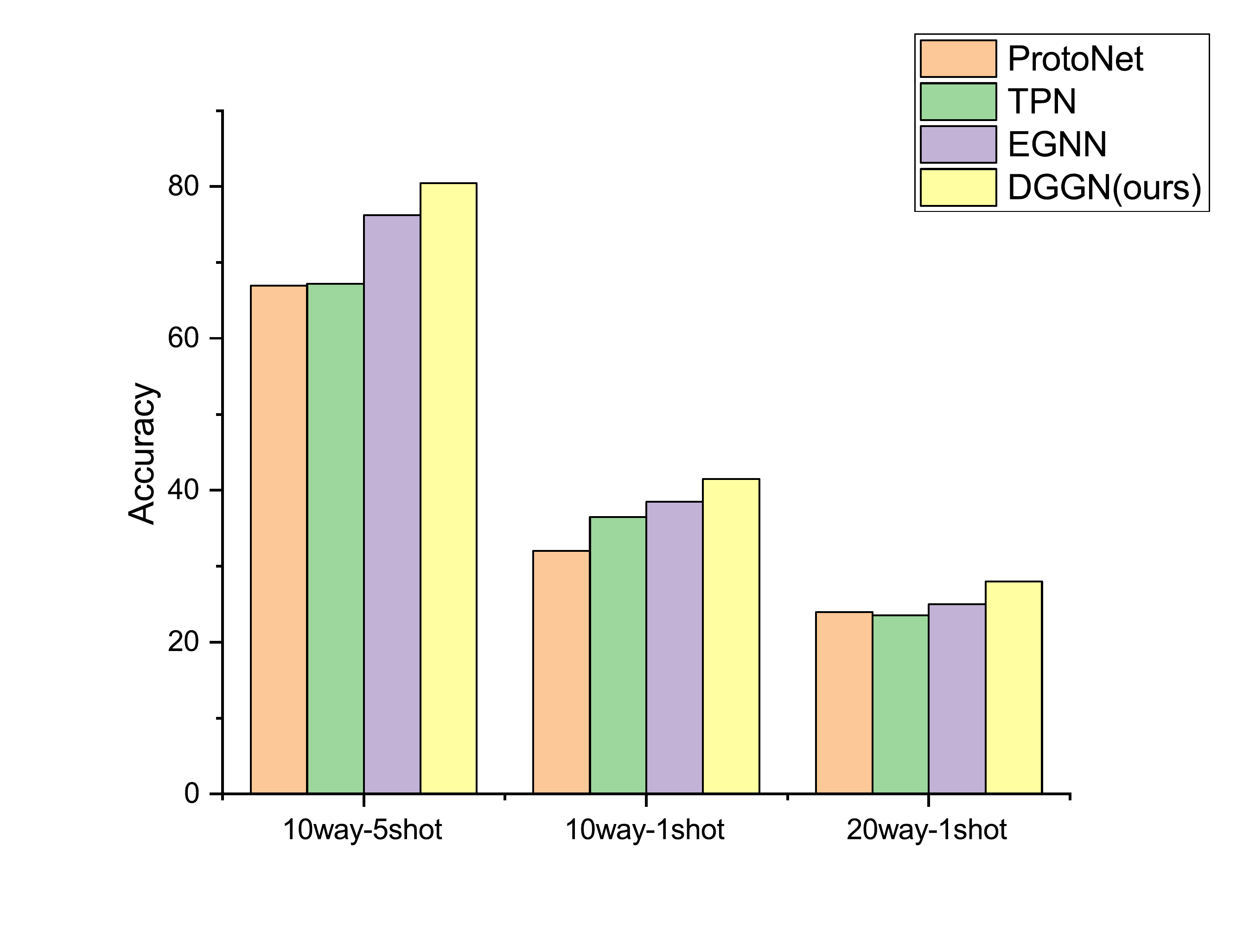}}
\end{minipage}
\caption{High-way few-shot classification performance on miniImagenet. }
\label{fig:highway}
\end{figure}

\section{CONCLUSION}
\label{sec:typestyle}

This paper describe a DGGN model containing a gated node aggregation module and an improved GRU based edge update module for a few-shot classification. The node aggregation module adopts a gate mechanism based on activation of the edge feature, making a learnable node update process. On the other hand, the edge update module employs an improved GRU unit, where the gradient backpropagation benefit from the GRU sequence throughout all layers. It is also suitable for the GRU sequence to process the vector form node feature. Besides, the entire network is trained end-to-end. The experimental results show that our DGGN achieves a comparable performance to the-state-of-art methods.


%
%
%


\bibliographystyle{IEEEbib}
\bibliography{refs}

\begin{thebibliography}{10}

\bibitem{Krizhevsky2017ImageNet}
Krizhevsky, Alex, Sutskever, Ilya, Hinton, and E.~Geoffrey,
\newblock ``Imagenet classification with deep convolutional neural networks.,''
\newblock {\em Communications of the ACM}, 2017.

\bibitem{2017Revisiting}
Chen Sun, Abhinav Shrivastava, Saurabh Singh, and Abhinav Gupta,
\newblock ``Revisiting unreasonable effectiveness of data in deep learning
  era,''
\newblock in {\em 2017 IEEE International Conference on Computer Vision
  (ICCV)}, 2017.

\bibitem{2006One}
L~Feifei, R~Fergus, and P~Perona,
\newblock ``One-shot learning of object categories,''
\newblock {\em IEEE Trans Pattern Anal Mach Intell}, vol. 28, no. 4, pp.
  594--611, 2006.

\bibitem{lake2011one}
Brenden Lake, Ruslan Salakhutdinov, Jason Gross, and Joshua Tenenbaum,
\newblock ``One shot learning of simple visual concepts,''
\newblock in {\em Proceedings of the annual meeting of the cognitive science
  society}, 2011, vol.~33.

\bibitem{vinyals2016matching}
Oriol Vinyals, Charles Blundell, Timothy Lillicrap, Koray Kavukcuoglu, and Daan
  Wierstra,
\newblock ``Matching networks for one shot learning,''
\newblock {\em arXiv preprint arXiv:1606.04080}, 2016.

\bibitem{snell2017prototypical}
Jake Snell, Kevin Swersky, and Richard~S Zemel,
\newblock ``Prototypical networks for few-shot learning,''
\newblock {\em arXiv preprint arXiv:1703.05175}, 2017.

\bibitem{sung2018learning}
Flood Sung, Yongxin Yang, Li~Zhang, Tao Xiang, Philip~HS Torr, and Timothy~M
  Hospedales,
\newblock ``Learning to compare: Relation network for few-shot learning,''
\newblock in {\em Proceedings of the IEEE conference on computer vision and
  pattern recognition}, 2018, pp. 1199--1208.

\bibitem{finn2017model}
Chelsea Finn, Pieter Abbeel, and Sergey Levine,
\newblock ``Model-agnostic meta-learning for fast adaptation of deep
  networks,''
\newblock in {\em International Conference on Machine Learning}. PMLR, 2017,
  pp. 1126--1135.

\bibitem{nichol2018first}
Alex Nichol, Joshua Achiam, and John Schulman,
\newblock ``On first-order meta-learning algorithms,''
\newblock {\em arXiv preprint arXiv:1803.02999}, 2018.

\bibitem{scarselli2008graph}
Franco Scarselli, Marco Gori, Ah~Chung Tsoi, Markus Hagenbuchner, and Gabriele
  Monfardini,
\newblock ``The graph neural network model,''
\newblock {\em IEEE transactions on neural networks}, vol. 20, no. 1, pp.
  61--80, 2008.

\bibitem{kipf2016semi}
Thomas~N Kipf and Max Welling,
\newblock ``Semi-supervised classification with graph convolutional networks,''
\newblock {\em arXiv preprint arXiv:1609.02907}, 2016.

\bibitem{garcia2017few}
Victor Garcia and Joan Bruna,
\newblock ``Few-shot learning with graph neural networks,''
\newblock {\em arXiv preprint arXiv:1711.04043}, 2017.

\bibitem{liu2018learning}
Yanbin Liu, Juho Lee, Minseop Park, Saehoon Kim, Eunho Yang, Sung~Ju Hwang, and
  Yi~Yang,
\newblock ``Learning to propagate labels: Transductive propagation network for
  few-shot learning,''
\newblock {\em arXiv preprint arXiv:1805.10002}, 2018.

\bibitem{kim2019edge}
Jongmin Kim, Taesup Kim, Sungwoong Kim, and Chang~D Yoo,
\newblock ``Edge-labeling graph neural network for few-shot learning,''
\newblock in {\em Proceedings of the IEEE/CVF Conference on Computer Vision and
  Pattern Recognition}, 2019, pp. 11--20.

\bibitem{chung2014empirical}
Junyoung Chung, Caglar Gulcehre, KyungHyun Cho, and Yoshua Bengio,
\newblock ``Empirical evaluation of gated recurrent neural networks on sequence
  modeling,''
\newblock {\em arXiv preprint arXiv:1412.3555}, 2014.

\bibitem{santoro2016meta}
Adam Santoro, Sergey Bartunov, Matthew Botvinick, Daan Wierstra, and Timothy
  Lillicrap,
\newblock ``Meta-learning with memory-augmented neural networks,''
\newblock in {\em International conference on machine learning}. PMLR, 2016,
  pp. 1842--1850.

\bibitem{ioffe2015batch}
Sergey Ioffe and Christian Szegedy,
\newblock ``Batch normalization: Accelerating deep network training by reducing
  internal covariate shift,''
\newblock in {\em International conference on machine learning}. PMLR, 2015,
  pp. 448--456.

\bibitem{marcheggiani2017encoding}
Diego Marcheggiani and Ivan Titov,
\newblock ``Encoding sentences with graph convolutional networks for semantic
  role labeling,''
\newblock {\em arXiv preprint arXiv:1703.04826}, 2017.

\bibitem{li2015gated}
Yujia Li, Daniel Tarlow, Marc Brockschmidt, and Richard Zemel,
\newblock ``Gated graph sequence neural networks,''
\newblock {\em arXiv preprint arXiv:1511.05493}, 2015.

\bibitem{he2016deep}
Kaiming He, Xiangyu Zhang, Shaoqing Ren, and Jian Sun,
\newblock ``Deep residual learning for image recognition,''
\newblock in {\em Proceedings of the IEEE conference on computer vision and
  pattern recognition}, 2016, pp. 770--778.

\bibitem{ren2018meta}
Mengye Ren, Eleni Triantafillou, Sachin Ravi, Jake Snell, Kevin Swersky,
  Joshua~B Tenenbaum, Hugo Larochelle, and Richard~S Zemel,
\newblock ``Meta-learning for semi-supervised few-shot classification,''
\newblock {\em arXiv preprint arXiv:1803.00676}, 2018.

\bibitem{russakovsky2015imagenet}
Olga Russakovsky, Jia Deng, Hao Su, Jonathan Krause, Sanjeev Satheesh, Sean Ma,
  Zhiheng Huang, Andrej Karpathy, Aditya Khosla, Michael Bernstein, et~al.,
\newblock ``Imagenet large scale visual recognition challenge,''
\newblock {\em International journal of computer vision}, vol. 115, no. 3, pp.
  211--252, 2015.

\bibitem{paszke2017automatic}
Adam Paszke, Sam Gross, Soumith Chintala, Gregory Chanan, Edward Yang, Zachary
  DeVito, Zeming Lin, Alban Desmaison, Luca Antiga, and Adam Lerer,
\newblock ``Automatic differentiation in pytorch,''
\newblock 2017.

\end{thebibliography}
\end{document}